\documentclass[sigconf]{acmart} 

\AtBeginDocument{%
  }

\copyrightyear{2022}
\acmYear{2022}
\setcopyright{rightsretained}

\makeatletter
\renewcommand{\@copyrightowner}{Johannes Haug | ACM. This is the author's version of the work. It is posted here for your personal use. Not for redistribution. The definitive Version of Record was published by ACM under the following DOI:}
\makeatother

\acmConference[CIKM '22]{Proceedings of the 31st ACM
International Conference on Information and Knowledge Management}{October
17--21, 2022}{Atlanta, GA, USA}
\acmBooktitle{Proceedings of the 31st ACM International Conference on Information
and Knowledge Management (CIKM '22), October 17--21, 2022, Atlanta, GA, USA}
\acmPrice{15.00}
\acmDOI{10.1145/3511808.3557257}
\acmISBN{978-1-4503-9236-5/22/10}

\usepackage{algorithm}
\usepackage{algorithmic}
\usepackage{xcolor}
\usepackage{pifont}
\usepackage{subfig}

\newcommand{\E}{\mathbb{E}}

\usepackage{bm}

\DeclareMathOperator*{\argmax}{arg\,max}

\bmdefine{\bx}{x}
\bmdefine{\bphi}{\phi}
\bmdefine{\bPhi}{\Phi}

\newtheorem{definition}{Definition}

\begin{document}

\title{Change Detection for Local Explainability in Evolving Data Streams}

\author{Johannes Haug}
\email{johannes-christian.haug@uni-tuebingen.de}
\affiliation{
  \institution{University of Tuebingen}
  \city{Tuebingen}
  \country{Germany}
}
\author{Alexander Braun}
\email{al.braun@student.uni-tuebingen.de}
\affiliation{
  \institution{University of Tuebingen}
  \city{Tuebingen}
  \country{Germany}
}
\author{Stefan Zürn}
\email{stefan.zuern@student.uni-tuebingen.de}
\affiliation{
  \institution{University of Tuebingen}
  \city{Tuebingen}
  \country{Germany}
}
\author{Gjergji Kasneci}
\email{gjergji.kasneci@uni-tuebingen.de}
\affiliation{
  \institution{University of Tuebingen}
  \city{Tuebingen}
  \country{Germany}
}
\renewcommand{\shortauthors}{Johannes Haug, Alexander Braun, Stefan Zürn, \& Gjergji Kasneci}

\begin{abstract}
As complex machine learning models are increasingly used in sensitive applications like banking, trading or credit scoring, there is a growing demand for reliable explanation mechanisms. Local feature attribution methods have become a popular technique for post-hoc and model-agnostic explanations. However, attribution methods typically assume a stationary environment in which the predictive model has been trained and remains stable. As a result, it is often unclear how local attributions behave in realistic, constantly evolving settings such as streaming and online applications. In this paper, we discuss the impact of temporal change on local feature attributions. In particular, we show that local attributions can become obsolete each time the predictive model is updated or concept drift alters the data generating distribution. Consequently, local feature attributions in data streams provide high explanatory power only when combined with a mechanism that allows us to detect and respond to local changes over time. To this end, we present CDLEEDS, a flexible and model-agnostic framework for detecting local change and concept drift. CDLEEDS serves as an intuitive extension of attribution-based explanation techniques to identify outdated local attributions and enable more targeted recalculations. In experiments, we also show that the proposed framework can reliably detect both local and global concept drift. Accordingly, our work contributes to a more meaningful and robust explainability in online machine learning.
\end{abstract}




\maketitle

\section{Introduction}
\begin{figure*}[t]
\centering
\captionsetup{width=.23\linewidth} 
\subfloat[t=1: The classes are separated with an accuracy $\approx$ 85\%.]{
    \includegraphics[width=.24\linewidth]{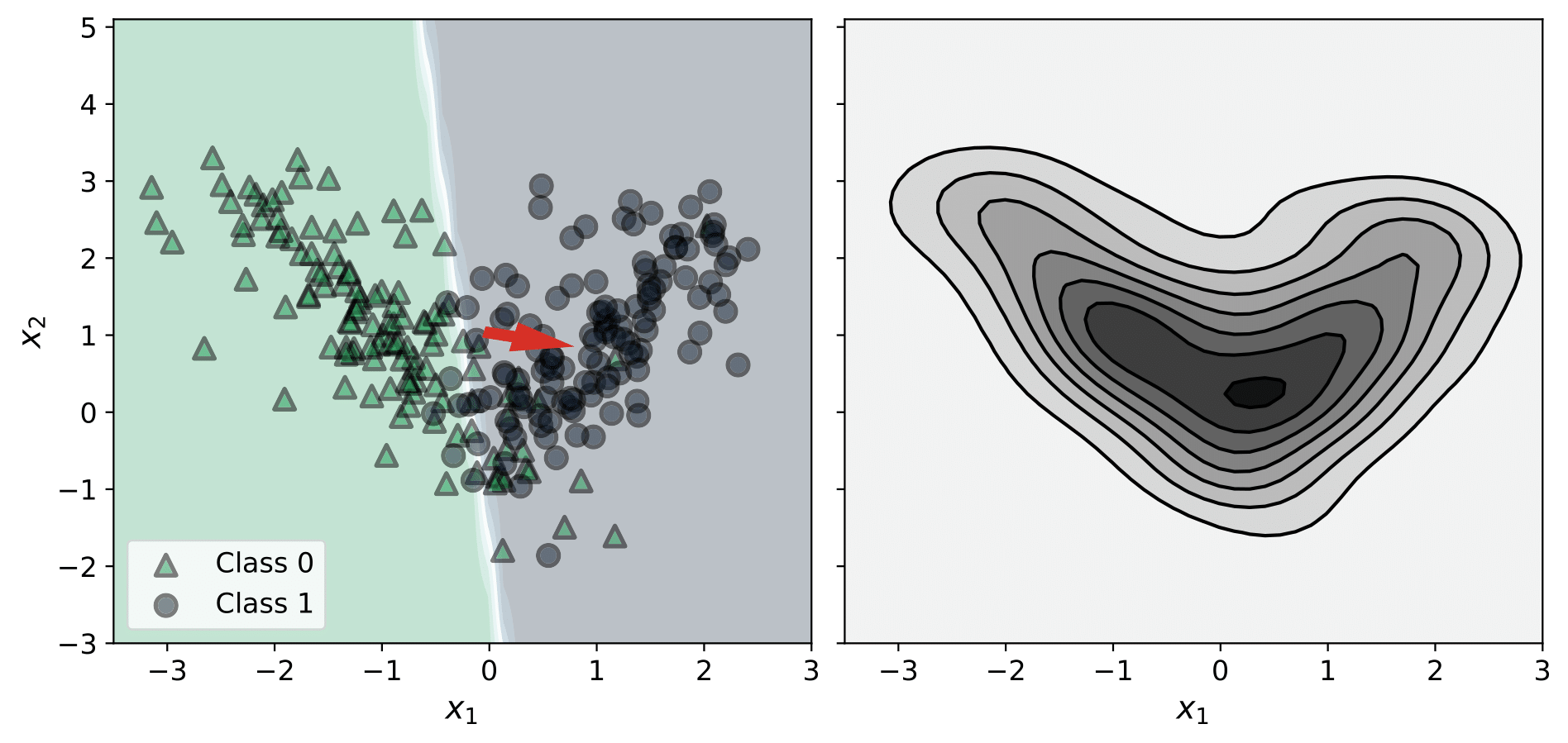}
}
\hfill
\subfloat[t=2: Virtual concept drift of the positive class (the conditional target distribution remains unaffected).]{
    \includegraphics[width=.24\linewidth]{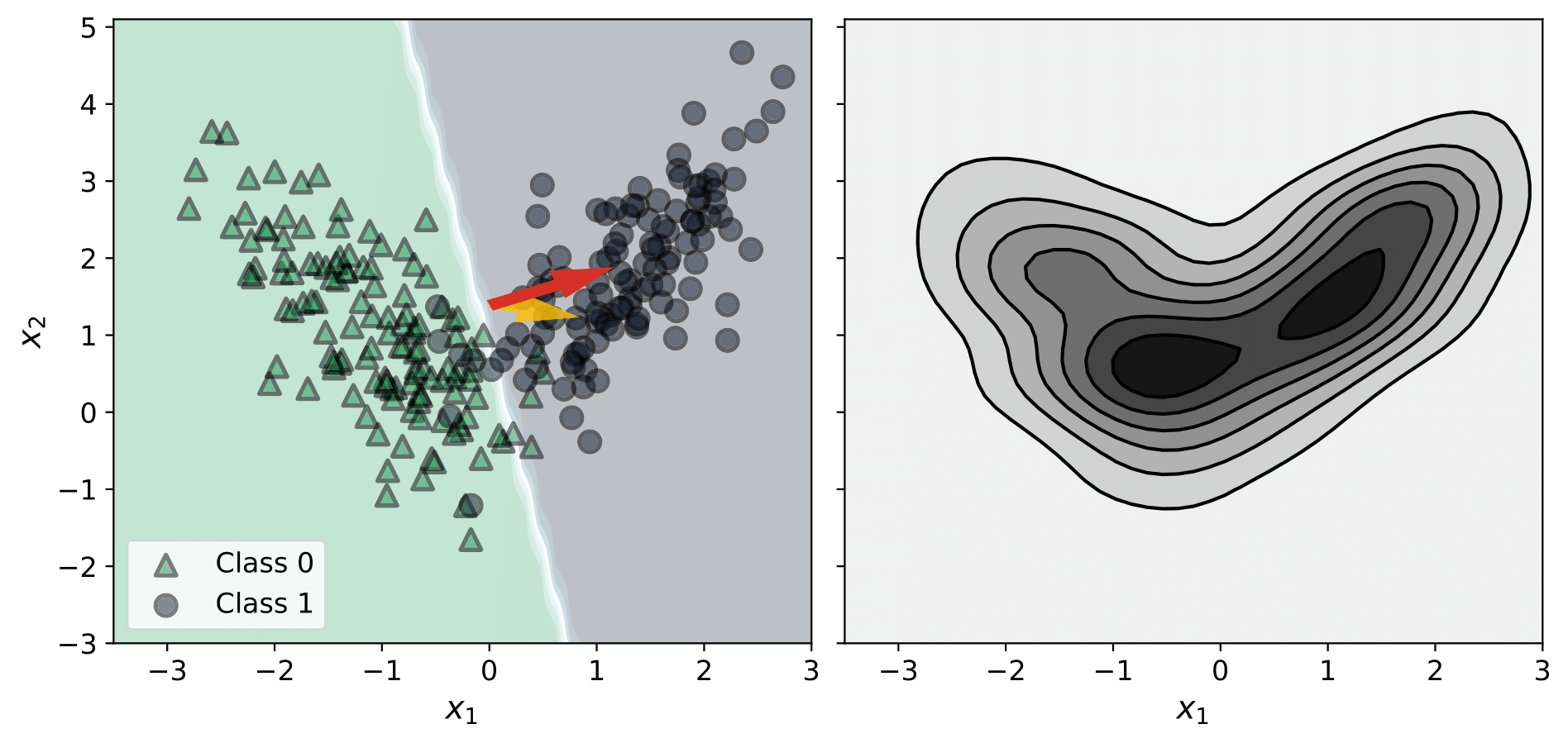}
}
\hfill
\subfloat[t=3: Real concept drift (the conditional target distribution has changed).]{
    \includegraphics[width=.24\linewidth]{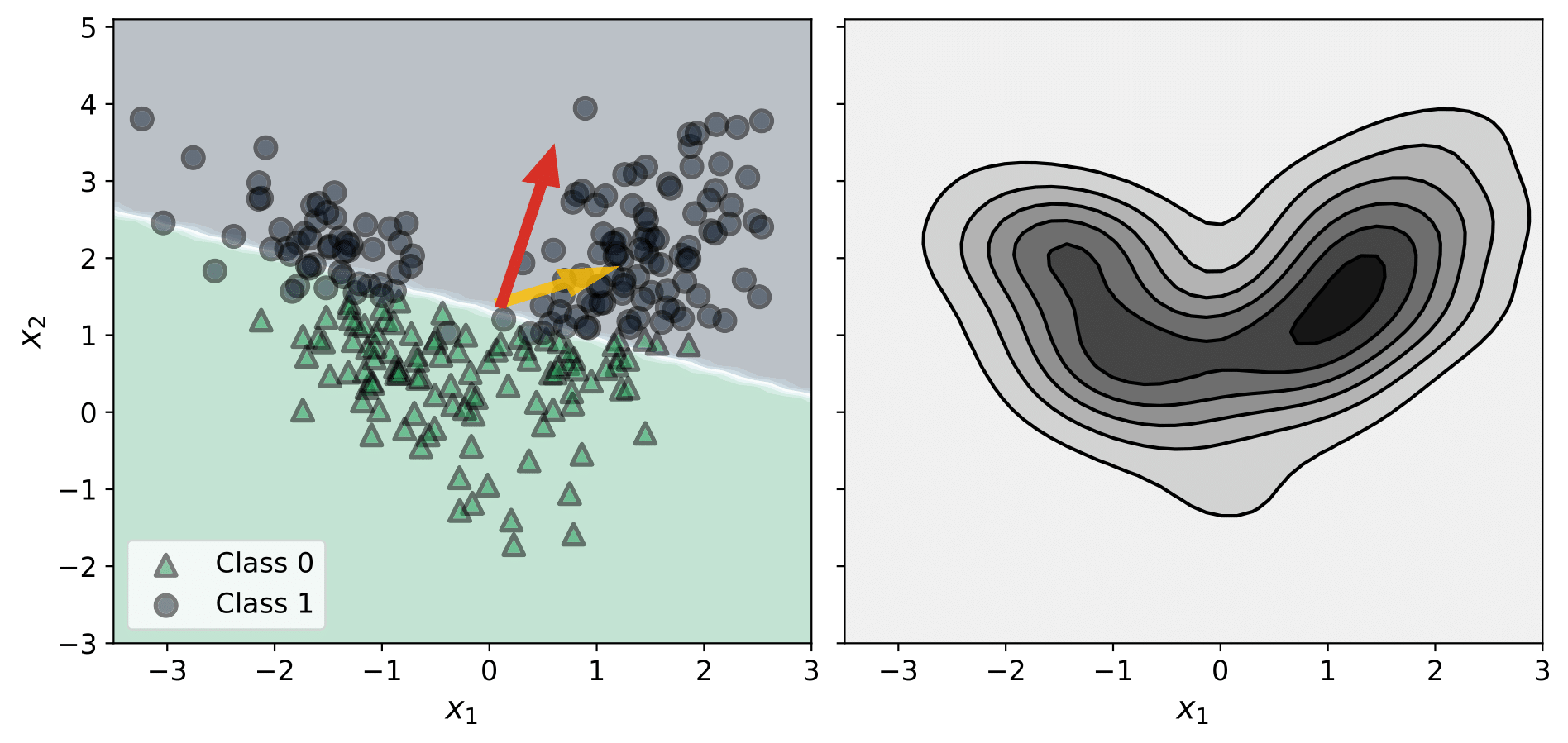}
}
\hfill
\subfloat[t=4: Virtual and real concept drift.]{
    \includegraphics[width=.24\linewidth]{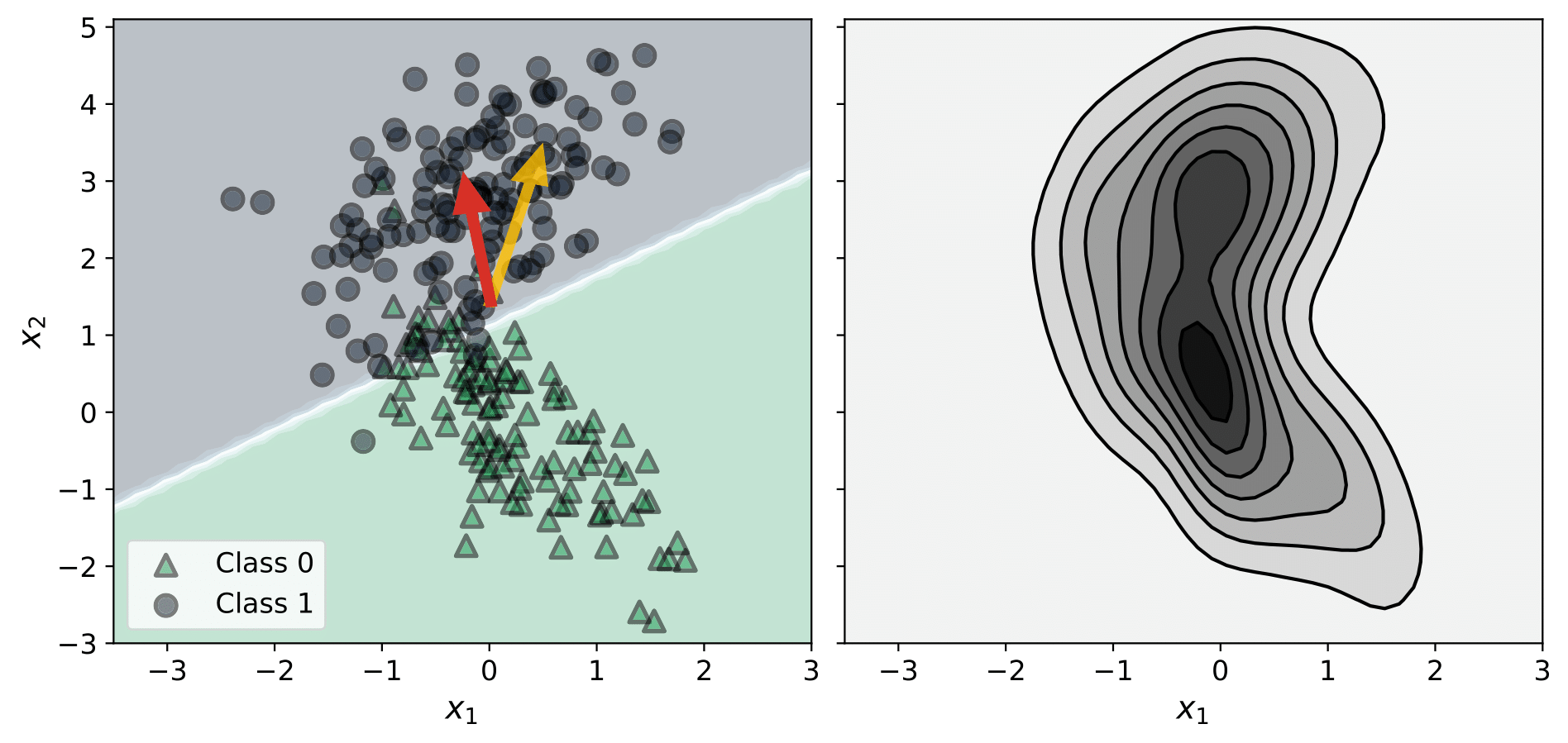}
}
\captionsetup{width=\linewidth} 
\caption{\textit{Local Attributions in a Synthetic Data Stream.} Concept drift can cause drastic changes of the local feature attributions obtained in a streaming application, which we illustrate above with synthetic data. We used a logistic regression classifier and the SHAP attribution framework \cite{lundberg2017unified}. Each plot depicts one of four time steps. The left subplots show the current data batch (250 observations), the decision boundary (between the green and blue areas) and the mean SHAP attribution of the current (red arrow) and previous time step (yellow arrow). The right subplots show the kernel density estimate of the observations. \textit{Both real and virtual concept drift change the decision boundary and thus the expected attribution between time steps.}}
\label{fig:toy_example}
\end{figure*}

Data streams are abundant in modern applications such as financial trading, social media, online retail, sensor-driven production or urban infrastructure \cite{gama2009overview}. To perform machine learning on large amounts of streaming data, we require powerful and efficient online learning models. Likewise, if we are to use online machine learning for high-stakes decisions, e.g. in online credit scoring or healthcare \cite{ta2016big}, we need reliable mechanisms to explain the model and its predictions. However, the explainability of online machine learning models has received only little attention in the past.

Online machine learning is generally more challenging than its offline counterpart. Aside from limited resources and real-time demands, online learning models must deal with changing environments and, in particular, concept drift, i.e., a shift in the data generating distribution \cite{haug2022dynamic}. Concept drift can manifest itself in most practical applications. For example, an online retailer must adapt product recommendations to changing customer preferences. Similarly, social media platforms need to consider the shifting interests of their users. If we do not account for concept drift, the performance and reliability of online learning methods can suffer. 

Due to stricter regulations and increased public awareness, interest in mechanisms for explainable machine learning has gained momentum in recent years. In this context, local feature attribution methods have become one of the most popular families of post-hoc explanation models \cite{ribeiro2016should,kasneci2016licon,lundberg2017unified,plumb2018model}. Local attribution methods aim to quantify the local importance of input features in the prediction. Traditionally, local attribution methods are used to explain the complex predictive model once it is trained and stationary. However, in data streams, concept drift requires that we continue updating the predictive model; accordingly, its explanation must also be updated. For example, we need to ensure that the explanations we give to a credit applicant are still meaningful after we update the predictive model with new customer data. However, although local attribution methods are commonly used, it is usually unclear how they behave in a realistic and dynamic online environment.

\begin{figure}[t]
\centering
\includegraphics[width=\linewidth]{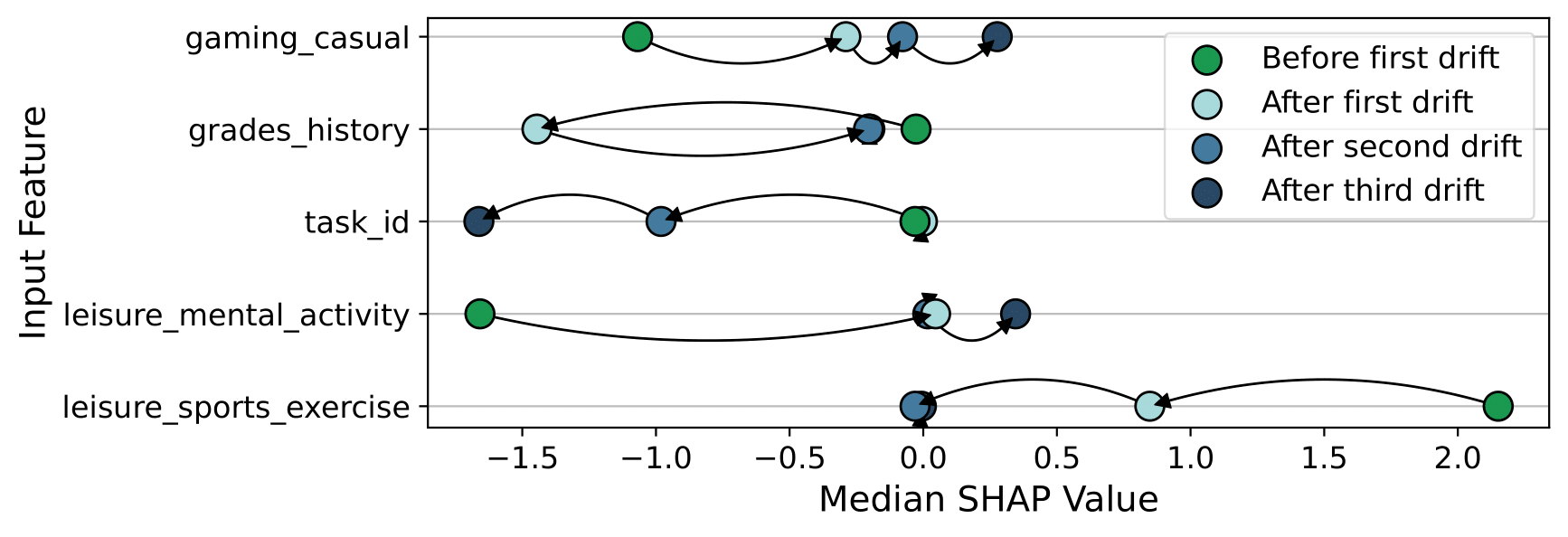}
\caption{\textit{Local Attributions in a Real-World Setting.} We trained a logistic regression classifier on the TüEyeQ data set \cite{kasneci2021tueyeq}, which comprises sociodemographic information of 315 subjects in an IQ test. The data set contains natural concept drifts by switching between 4 blocks of IQ-related tasks. Above, we show the median SHAP attributions \cite{lundberg2017unified} of the 5 input features with the highest variation over time. Online training and concept drift lead to drastic changes in the attributions of this real-world stream of tasks (i.e., solving IQ test items). For example, the \textit{task-id} has a greater (negative) importance in later task blocks that are more difficult to solve (for more information about the features see \cite{kasneci2021tueyeq}). \textit{To achieve better feature-based explanations in online scenarios, we need to identify such (local) changes of the attributions.}}
\label{fig:shap_iq_change}
\end{figure}

\subsection{Local Attributions Under Concept Drift}
Indeed, a simple example shows that local attributions for an incrementally trained machine learning model can change significantly over time. For illustration, we generated an artificial two-dimensional data set that underlies different types of concept drift (we discuss the two fundamental types of concept drift more formally in Section \ref{sec:preliminaries}). Figure \ref{fig:toy_example} illustrates four time steps from the training procedure of a logistic regression model. Strikingly, the expected feature attributions (SHAP values \cite{lundberg2017unified}, red and yellow arrows) changed drastically due to shifts in the decision boundary of the classifier (green and blue areas), which in turn were caused by concept drift. This example shows that local attributions generated at a certain point in time may lose their validity after a single update of the predictive model. In fact, we observe such behavior in real-world settings. For example, the SHAP attributions of a recent IQ study \cite{kasneci2021tueyeq} underlie considerable change over time (Figure \ref{fig:shap_iq_change}). For the long-term acceptance of machine learning models in sensitive applications, one has to address such changes in the data.


Ideally, local feature attribution methods should account for changes in a data stream by design, e.g., through robustness to small model variations or efficient incremental update procedures. However, most existing attribution methods produce point estimates and rely on costly sampling- or permutation-based approximations \cite{lundberg2017unified,ribeiro2016should}. Thus, without knowledge of the hidden dynamics underlying a data stream -- and in particular concept drift -- we would have to recalculate the local attributions over and over again to ensure their validity. Given the scale of most data streams, however, it is typically infeasible to simply recompute all past attribution vectors at every time step. In fact, since incremental updates, e.g., via stochastic gradient descent, often only alter parts of the model, recalculation of all previous attributions would often be unnecessary. Therefore, in order to efficiently establish local feature attributions in data streams, enable more informed decisions, and provide an overall higher degree of explainability, we require an effective mechanism to detect local changes over time. Indeed, we argue that \textit{(local) change detection should be part of any sensible strategy for explainable machine learning in evolving data streams.}

\subsection{Our Contribution}
In this paper, we introduce a novel framework for \textit{Change Detection for Local Explainability in Evolving Data Streams (CDLEEDS)}. CDLEEDS serves as a generic and model-agnostic extension of local attribution methods in data streams. The proposed change detection strategy arises naturally from the behavior of local attribution methods in the presence of incremental model updates and concept drift. In fact, we show that due to a fundamental property of many attribution methods, local change can be reliably detected without computing a single attribution score. We propose an effective implementation of CDLEEDS via adaptive hierarchical clustering. In experiments, we show that our approach can help to significantly reduce the number of recalculations of a local attribution over time. 
CDLEEDS is also one of the first drift detection methods capable of detecting concept drift at different levels of granularity. Thus, CDLEEDS can have general value for online machine learning applications -- even outside the context of explainability. For illustration, we compared our model with several state-of-the-art drift detection methods. Notably, CDLEEDS was able to outperform existing methods for both real-world and synthetic data streams.

In summary, this paper is one of the first to discuss local explainability and, in particular, local feature attributions in evolving data streams. We propose a powerful and flexible framework capable of recognizing local and global changes in the online learning and attribution model. In this way, our work enables more meaningful and robust use of local feature attribution methods in data streams and is thus an important step towards better explainability in the practical domain of online machine learning.

In Section \ref{sec:related_work}, we introduce related work. In Section \ref{sec:preliminaries}, we formally examine the behavior of local attributions under concept drift. We then present CDLEEDS and an effective implementation in Section \ref{sec:cdleeds}. Finally, in Section \ref{sec:experiments}, we demonstrate CDLEEDS in several experiments on binary and multiclass tabular streaming data sets.

\section{Related Work}\label{sec:related_work}
In the following, we briefly outline related work on local attribution methods, explainability in data streams and concept drift detection. 

\paragraph{Local Feature Attributions}
Local attribution methods are one of the most popular and widely used explanation techniques. Most frameworks are based on a similar intuition: a complex (black-box) predictive model can be locally approximated sufficiently well by a much simpler, usually linear, explanation model. Model-agnostic frameworks like LIME \cite{ribeiro2016should} and SHAP \cite{lundberg2017unified} belong to the most popular attribution methods and have inspired a variety of follow-up work \cite{sundararajan2019many,chen2020true,aas2021explaining,jesus2021can,lundberg2018consistent}. Additionally, there are model-specific attribution methods that exploit the inner mechanics of the complex model. In particular, a large literature has formed on gradient-based attribution techniques for Deep Neural Networks \cite{kasneci2016licon,shrikumar2017learning,ancona2018towards,sundararajan2017axiomatic}. For more detailed information, we refer to recent surveys \cite{adadi2018peeking,guidotti2018survey,carvalho2019machine,arrieta2020explainable}.

\paragraph{Explainability in Data Streams}
Compared to the explanation of offline (black-box) models, relatively little attention has been paid to the explainability of predictions in dynamic data streams. \citet{bosnic2014enhancing} were among the first to describe that explanations in a data stream must actually consist of a series of individual explanations that can change over time. \citet{demvsar2018detecting} later argued that the dissimilarity of periodically generated feature attributions may be used to detect concept drift. Finally, \citet{tai2018sketching} briefly discuss feature-based explainability in the context of sketching. Still, given the abundance of streaming applications in practice, explainable machine learning for data streams should receive more attention.

\paragraph{Concept Drift Detection}
Concept drift detection has traditionally served as a tool to prevent deterioration in predictor performance following distributional changes. As such, global concept drift detection methods have been integrated into state-of-the-art online learning frameworks, such as the Hoeffding Tree \cite{bifet2009adaptive}. Most modern drift detection methods are based on changes in the observed predictive error of the online learning model \cite{gama2004learning,baena2006early,barros2017rddm}. In this context, many approaches use sliding windows for more robust or statistically significant drift detection \cite{pesaranghader2016fast,bifet2007learning,ross2012exponentially}. Alternatively, a more recent approach monitors changes in the inherent uncertainty of model parameters to detect global and feature-specific (partial) concept drift \cite{haug2021learning}. For more information about global concept drift detection, we refer to the comprehensive surveys of \citet{zliobate2010learning,gama2014survey,webb2016characterizing,gonccalves2014comparative}.

In contrast, the potential explanatory power of concept drift detection has been largely neglected. Accordingly, there are only few methods capable of local concept drift detection. For example, \citet{gama2006learning} integrate an error-based concept drift detection scheme into the inner nodes of an incremental decision tree. In this way, they are able to detect concept drifts in specific input regions, represented by the branches of the tree. However, as mentioned earlier, we need a mechanism that is able to detect instance-level changes to enable better explainability in a data stream.

\section{Online Learning and Local Attribution -- Formal Introduction}
\label{sec:preliminaries}
Data streams are defined by a series of time steps $1,\dots,t,\dots,T$. At each time step, we obtain an observation $x_t \in \mathbb{R}^m$ and a corresponding label $y_t \in \mathbb{R}$, where $m$ is the number of features. Our goal is to incrementally train an online predictive model $f_{\theta_t}(x_t)$. That is, we aim to optimize the model parameters $\theta_t$ at every time step $t$ given the new training observation (we may also use batches of training data). Since the parameters $\theta_t$ are defined by the selected model, we write $f_{\theta_t}(x_t) = f_t(x)$ to simplify the exposition.

We may represent the observations and labels by two corresponding random variables $X$ and $Y$. The data generating concept at time step $t$ is defined by the joint probability distribution $P_t(Y,X)$. Typically, we assume that the observations are drawn independently from the data generating distribution. Although this independence assumption can be violated in practice, it has proven effective in many applications \cite{haug2020leveraging}. Concept drift describes a change of the joint probability between two time steps $t_1$ and $t_2$ \cite{webb2016characterizing}, i.e.,
\begin{equation}\label{eq:def_concept_drift}
    P_{t_1}(Y,X) \neq P_{t_2}(Y,X) \Leftrightarrow P_{t_1}(Y|X) P_{t_1}(X) \neq P_{t_2}(Y|X) P_{t_2}(X).
\end{equation}
In general, we distinguish two fundamental types of concept drift. \textit{Real} concept drift corresponds to a change in the conditional probability distribution $P_t(Y|X)$, while $P_t(X)$ remains stable. Conversely, \textit{virtual} concept drift describes a shift in $P_t(X)$, while $P_t(Y|X)$ remains unchanged. Other than real concept drift, virtual concept drift does not change the optimal decision boundary. In practice, we are therefore mostly interested in real concept drift, i.e. $P_{t_1}(Y|X) \neq P_{t_2}(Y|X)$. Nevertheless, virtual concept drift may affect the (learned) decision boundary of our online learning model \cite{oliveira2021tackling}. This effect can be seen in our introductory example in Figure \ref{fig:toy_example}b. Moreover, we can distinguish between local and global concept drift. While global concept drift affects the entire (or large regions) of the input space, local concept drift is locally bounded. Hence, it is often more difficult to detect local concept drift.

In practice, the true data generating distribution is usually unknown. Hence, the online predictive model is often our best approximation of the active concept \cite{haug2021learning}. That is, we typically assume that the predictive model at time step $t$ approximates the conditional target probability well, i.e. $P_t(Y|X) \approx f_t(x)$. This simplifying assumption is the fundamental basis of most existing concept drift detection methods \cite{haug2021learning,bifet2007learning,baena2006early}. Therefore, instead of explicitly learning the true data generating distribution, we can detect concept drift directly from a change in the predictive model:
\begin{equation}\label{eq:approx_concept_drift}
    f_{t_1}(x) \neq f_{t_2}(x)
\end{equation}
Notably, since Eq. \eqref{eq:approx_concept_drift} allows us to detect concept drift based on changes in the decision boundary, we can also detect the changes caused by virtual concept drift as described above.

\subsection{Local Attribution Accuracy and Its Implication for Online Learning}
Local attribution methods allow us to explain complex predictive models by quantifying the local importance of input features in the prediction. Let $\phi_{x_i,f} \in \mathbb{R}^m$ be the local attribution vector corresponding to an observation $x_i$ and a model $f$. Typically, the generated feature attribution vector $\phi_{x_i,f}$ adheres to a set of sensible properties. A fundamental property shared by most state-of-the-art attribution methods is \textit{local accuracy} \cite{lundberg2017unified}, also known as \textit{local fidelity} \cite{ribeiro2016should} or \textit{summation to delta} \cite{shrikumar2017learning}. 

Local accuracy describes that the attribution vector must account for the difference between the local model outcome and a baseline value. We can adopt the generic definition of \citet{lundberg2017unified} for the online case and define local accuracy accordingly as
\begin{equation}\label{eq:local_accuracy}
    f_t(x_i) = \phi^0_t + \sum^m_{j=1} \phi^j_{x_i,f_t},
\end{equation}
where $\phi^0_t \in \mathbb{R}$ is the baseline outcome at time step $t$ and $\phi^j_{x_i,f_t}$ is the attribution of feature $j$. Note that \citet{lundberg2017unified} used an additional vector representation of missing features. In general, however, we can assume that the observation to be explained has no missing features.

The baseline value $\phi^0_t$ is set to represent missing discriminative information, i.e., ideally it is a value for which the prediction is neutral. For example, in image recognition, the zero vector is often used as a baseline \cite{sundararajan2017axiomatic}. Alternatively, we might use the expectation $\phi^0 = \E_{X^0}[f(x)]$ over a static sample of training observations $X^0$ as our baseline \cite{lundberg2017unified}. The choice of the baseline can drastically alter the generated attributions and should thus be selected with care \cite{haug2021baselines}. In particular, for data streams where our understanding of missingness may change over time, we might need to update $\phi^0_t$ between time steps. We propose an effective baseline in Section \ref{sec:algorithmic_decisions}.

In the introductory experiments, we have shown that local feature attributions may lose their validity due to incremental model updates and concept drift. With the above definitions in place, we can now express this behavior in more formal terms. Suppose the predictive model has changed between two time steps $t_1$ and $t_2$ according to \eqref{eq:approx_concept_drift}. We know that there must exist at least one data point $x_i$ such that $f_{t_1}(x_i) \neq f_{t_2}(x_i)$. By definition of local accuracy, a shift of the local model outcome $f_t(x_i)$ implies a shift of the baseline $\phi^0_t$ and/or the local attribution vector $\phi_{x_i,f_t}$ and vice versa:
\begin{equation}\label{eq:attribution_diff}
    f_{t_1}(x_i) \neq f_{t_2}(x_i) \overset{\eqref{eq:local_accuracy}}{\Leftrightarrow} \phi^0_{t_1} + \sum^m_{j=1} \phi^j_{x_i,f_{t_1}} \neq \phi^0_{t_2} + \sum^m_{j=1} \phi^j_{x_i,f_{t_2}}
\end{equation}
In other words, any change in the decision boundary of the model, e.g., due to concept drift or incremental updates, is guaranteed to change either the baseline and/or at least one local attribution score. Therefore, as before, we argue that local attribution methods in data streams need a mechanism to detect such local changes in order to provide meaningful explanations over time.

\section{The CDLEEDS Framework}\label{sec:cdleeds}
We present CDLEEDS, a novel framework for local change detection that allows us to identify outdated attributions and enable more efficient and targeted recalculations for temporally adjusted explanations in data streams. In general, our goal is to identify whether a local attribution vector $\phi_{x_i,f_t}$ has changed between two time steps $t_1$ and $t_2$. Based on the local accuracy property, we can immediately formulate a na\"ive scheme for local change detection:
\begin{equation}\label{eq:basic_detection_scheme}
    \sum^m_{j=1} \phi^j_{x_i,f_{t_1}} \neq \sum^m_{j=1} \phi^j_{x_i,f_{t_2}} \overset{\eqref{eq:local_accuracy}}{\Leftrightarrow} f_{t_1}(x_i) - \phi^0_{t_1} \neq f_{t_2}(x_i) - \phi^0_{t_2}.
\end{equation}
By calculating the right part of Eq. \eqref{eq:basic_detection_scheme} for a given observation $x_i$ in all successive time steps, we are able to detect local change over time. Indeed, since we only require the baseline $\phi^0_t$ and model outcome $f_t(x_i)$, \textit{this simple method allows us to detect local changes without calculating a single attribution vector.} However, this approach may be too costly if we want to detect changes for a large number of observations (because we would have to repeatedly obtain predictions $f_t(x_i)$, e.g., for an entire user base). Moreover, since we are comparing snapshots at individual time steps, this na\"ive approach might be prone to noise. Therefore, we need to modify this basic change detection method to make it more reliable and efficient.

\subsection{Spatiotemporal Neighborhoods}\label{sec:spatiotemporal_neighborhood}
In practice, it may often be sufficient to detect changes in the close proximity of a given observation. Specifically, if we can detect concept drift in the neighborhood of an observation $x_i$ with high confidence, it is likely that the attribution of $x_i$ has changed. To this end, we need a meaningful understanding of neighborhood in data streams. Intuitively, we would like a neighborhood to include close previous observations. In this context, we introduce the notion of \textit{spatiotemporal neighborhood:}
\begin{definition}[Spatiotemporal $\gamma$-Neighborhood (STN)]\label{def:stn}
    Let $\text{sim}(\cdot)$ be a sensible similarity measure (e.g., cosine similarity or RBF kernel). A spatiotemporal $\gamma$-neighbourhood with respect to an observation $x_i$ is defined by a set of time steps $\Omega^{(x_i,\gamma)} = \{t \in \{1,\dots,T\}~|~\text{sim}(x_t, x_i) \geq \gamma\}$.
\end{definition}
More intuitively, a spatiotemporal $\gamma$-neighborhood, STN for short, is a set of time steps corresponding to previous observations similar to the observation in question. With the parameter $\gamma$ we can control the minimal similarity and thus the boundedness of the STN. If we are able to detect changes in the STN of an observation $x_i$ with reasonably large $\gamma$, we can assume that the attribution of that observation has changed. Accordingly, we can rephrase the na\"ive change detection method from Eq. \eqref{eq:basic_detection_scheme} in a more robust way:
\begin{equation}\label{eq:drift_detection_method}
    \E_{u \in \Omega^{(x_i,\gamma)}_{<t}}[f_u(x_u) - \phi^0_u] \neq \E_{v \in \Omega^{(x_i,\gamma)}_{\geq t}}[f_v(x_v) - \phi^0_v],
\end{equation}
where $\Omega^{(x_i,\gamma)}_{<t} = \{u \in \{1,\dots,t-1\} ~|~ \text{sim}(x_u, x_i) \geq \gamma\}$ and $\Omega^{(x_i,\gamma)}_{\geq t} = \{v \in \{t,\dots,T\} ~|~ \text{sim}(x_v, x_i) \geq \gamma\}$ denote the STNs of $x_i$ for different intervals before and after the time step $t$. With Eq. \eqref{eq:drift_detection_method}, we can now compare time intervals instead of individual snapshots, which usually leads to more robust and reliable detections. Note that we can scale the time intervals, and hence the size of the STNs, by limiting the set of relevant time steps. For example, to obtain the STN in an interval of size $w$ before time step $t$, we can specify $\Omega^{(x_i,\gamma)}_{<t} = \{u \in \{t-w,\dots,t-1\} ~|~ \text{sim}(x_u, x_i) \geq \gamma\}$. Similar to existing concept drift detection methods that use sliding windows (see Section \ref{sec:related_work}), the size of the specified time intervals affects the performance. If the interval is chosen too small, the method may not be robust and produce false alarms. On the other hand, if the interval is chosen too large, certain changes may be missed. In order to achieve a higher degree of flexibility, we therefore only limit the maximum size of an STN in our implementation, but not the eligible time intervals.

In order to detect local change over time, we can incrementally update the STNs $\Omega^{(x_i,\gamma)}_{<t}$ and $\Omega^{(x_i,\gamma)}_{\geq t}$. As a result, we avoid having to consider (predict) old observations repeatedly, which considerably reduces the resource consumption compared to the initial na\"ive scheme. In fact, since we can process observations in a single pass, we fulfill a central requirement of online machine learning \cite{domingos2001catching}.

Instead of comparing expectations directly, as shown in Eq. \eqref{eq:drift_detection_method}, we may also use a hypothesis test to detect significant changes over time. Note that we have assumed independent streaming observations (see Section \ref{sec:preliminaries}). Moreover, the expectations in Eq. \eqref{eq:drift_detection_method} tend to be normally distributed for large sample sizes, i.e., for large STNs, according to the central limit theorem. Therefore, if we specify reasonably large STNs, we may apply the unpaired two-sample t-test (which we did in our implementation).


\begin{algorithm}[t]
\caption{update() - General update procedure at a node $n$ of the CDLEEDS hierarchical clustering approach.}
\label{alg:pseudo_code}
\begin{algorithmic}[1]
\renewcommand{\algorithmiccomment}[1]{\bgroup\hfill//~#1\egroup}
\REQUIRE Observation $x_t$; Prediction-baseline difference $\hat{y}_t - \phi^0_t$.
\\ \textit{*** A node comprises an age counter, a sliding window of observations used for clustering, a sliding window of prediction-baseline differences used for change detection, and a centroid. ***}
\\ \textit{*** The sliding windows $W_n$, $V_n$ have a user-defined size and correspond to an STN at the centroid, i.e. $\Omega^{(c_n,\gamma)} = \{u \in \{t-w,\dots,t\}\ ~|~ \text{sim}(x_u,c_n) \geq \gamma\}$. ***}
\STATE $\text{age}_n \leftarrow \text{age}_n + 1$
\STATE $W_n \leftarrow$ Remove oldest entry and append $x_t$.
\STATE $V_n \leftarrow$ Remove oldest entry and append $\hat{y}_t - \phi^0_t$.
\STATE $c_n \leftarrow \text{mean}(W_n)$
\item[]
\IF {$n$ is a leaf node}
    \IF {$\exists x_u \in W_n: \text{sim}(x_u, c_n) < \gamma$}
        \STATE \textit{*** Split the node by using the most dissimilar points in $W_n$ as the centroids of the new children. ***}
        \STATE $n_{\text{left}}, n_{\text{right}} \leftarrow$ Split the node $n$.
        \STATE \textit{*** Assign each observation to the closest child node. ***}
        \FOR{$x_u \in W_n$}
            \STATE $n_{\text{child}} \leftarrow \underset{[n_{\text{left}},n_{\text{right}}]}{\argmax}\big(\text{sim}(x_u,c_{n_\text{left}}),\text{sim}(x_u,c_{n_\text{right}})\big)$
            \STATE $n_{\text{child}}.\text{update}(x_u, ~\hat{y}_u - \phi^0_u)$
            \STATE $\text{age}_{n_{\text{child}}} \leftarrow \text{age}_n$
        \ENDFOR
    \ELSE
        \STATE \textit{*** Identify change at the node by testing for a significant difference in $V_n$. According to Eq. \eqref{eq:drift_detection_method}, we compare the means of the first and second (equally sized) halves of $V_n$.***}
        \STATE $\bar{V}^*_n = \text{mean}(V_n[:|V_n|/2])$
        \STATE $\bar{V}^{**}_n = \text{mean}(V_n[|V_n|/2:])$
        \IF {$h_0: \bar{V}^*_n = \bar{V}^{**}_n$ can be rejected for significance $\alpha$}
            \STATE Alert local change at $n$.
        \ENDIF
    \ENDIF
\ELSE
    \STATE \textit{*** Forward $x_t$ to the closest child. ***}
    \STATE $n_{\text{child}} \leftarrow $ see line 11.
    \STATE $n_{\text{child}}.\text{update}(x_t, ~\hat{y}_t - \phi^0_t)$
    \STATE $\text{age}_{n_{\text{child}}} \leftarrow \text{age}_n$
    \\ \textit{*** Check if the split is outdated and should be pruned. ***}
    \IF {$\text{age}_n - \text{min}\big(\text{age}_{n_{\text{left}}}, ~\text{age}_{n_{\text{right}}}\big) \geq$ threshold}
        \STATE Prune the branch at $n$ and make $n$ a leaf node.
        \STATE Test for change at $n$ as in line 16 - 21.
    \ENDIF
\ENDIF
\end{algorithmic}
\end{algorithm}

\subsection{Finding Representative Neighborhoods With Adaptive Hierarchical Clustering}\label{sec:cluster}
Data streams produce a large number of observations for which we may need to detect changes in the explanation. Although Eq. \eqref{eq:drift_detection_method} provides an efficient mechanism for detecting changes at a point $x_i$, the construction of STNs for all observations to be explained can lead to a high computational cost. For practical reasons, we may instead select a representative set of observations for which we maintain STNs over time, which in turn serve as an approximation to the STNs of all observations. Specifically, since we are interested in grouping similar data points according to Definition \ref{def:stn}, we aim to identify a set of representative observations $C_t = \{c_1,\dots,c_n,\dots,c_N\}$, such that each $c_n$ is similar to a large group of current observations. This problem is very similar to online clustering \cite{cao2006density,zhang1996birch}, where each $c_n$ denotes the centroid of a cluster $\Gamma_{c_n} = \{x_i~|~ \text{sim}(x_i,c_n) \geq \gamma\}$. Accordingly, if we obtain an STN with respect to $c_n$ in a given interval, e.g. $\Omega^{(c_n, \gamma)}_{<t}$, we can assume that it is also representative of all observations in the cluster, and hence
\begin{equation}\label{eq:approx_leaf_stn}
    \forall x_i \in \Gamma_{c_n}: \E_{u \in \Omega^{(x_i,\gamma)}_{<t}}[f_u(x_u) - \phi^0_u] \approx \E_{u \in \Omega^{(c_n,\gamma)}_{<t}}[f_u(x_u) - \phi^0_u].
\end{equation}
Note that Eq. \eqref{eq:approx_leaf_stn} holds equivalently for $\Omega^{(c_n, \gamma)}_{\geq t}$.

On this basis, we propose a simple hierarchical and dynamic clustering of observations in a binary tree. The root of the clustering tree contains all observations from a specified interval, implemented as a sliding window. The centroid corresponds to the mean value of these observations. If the similarity radius of the current node is smaller than $\gamma$, we split the node by choosing the two most dissimilar data points as the new children (i.e., a binary split). The observations of the parent node are then assigned to the most similar child. We continue the procedure for the children recursively until the similarity radius for each leaf node is greater or equal $\gamma$. 

Virtual concept drift can shift high-density regions in the input space (see Figure \ref{fig:toy_example}). Therefore, the clustering should adjust accordingly. For this purpose, we maintain an internal age counter for each node, which is updated as soon as the node receives a new observation. If a child node has not received any observations for a while, its age differs from the age of the parent node, indicating an outdated split that can be pruned.

To identify change for a given observation, we then only need to retrieve the most similar leaf node of the current tree and test for change as specified in Eq. \eqref{eq:drift_detection_method} using the STNs of the corresponding centroid (see Eq. \eqref{eq:approx_leaf_stn}). The general procedure at a node of the tree is described in Algorithm \ref{alg:pseudo_code}. A corresponding implementation is available at \url{https://github.com/haugjo/cdleeds}.

The proposed hierarchical clustering provides clusters with increasing granularity. In the context of explainable online learning, this is an advantage as we are able to detect change at different hierarchies. For example, to detect global change, we can combine the test results of leaf nodes with Fisher's method \cite{fisher1992statistical}. In this context, we would correct the significance level $\alpha$ for multiple hypothesis testing, using the mean false discovery rate $\alpha_{\text{corr}} = \alpha(N+1)/(2N)$, where $N$ is the number of independent tests at the leaf nodes.

\subsection{Further Algorithmic Decisions}\label{sec:algorithmic_decisions}
The generic clustering method proposed above requires us to make choices during implementation. A central component of the clustering is the similarity measure. The cosine similarity and the (negative) Euclidean distance are commonly used to measure the similarity of vectors. However, the cosine similarity is a measure of orientation and does not take into account the magnitude of the input features, which are relevant for local attributions. Conversely, the Euclidean distance is very sensitive to the dimensionality and magnitude of a vector. This can make it difficult to establish a meaningful threshold $\gamma$ - especially since dimensionality and magnitude can change in practice due to concept drift. For this reason, we use the Radial Basis Function (RBF) kernel in our implementation (with variance parameter $1/m$, where $m$ is the number of features). The RBF kernel ranges from zero to one (when the vectors are equal) and is frequently used as a measure of similarity in machine learning. Due to the boundedness of the RBF kernel, it is generally much easier to specify and interpret the parameter $\gamma$.

Moreover, we use the exponentially weighted moving average to obtain our baseline, i.e., $\phi^0_t = f_t(\text{EWMA}_t)$ with $\text{EWMA}_t = \beta x_t + (1- \beta) \text{EWMA}_{t-1}$, where $\beta \in [0,1]$ is the decay factor. Compared to using a static sample of observations \cite{lundberg2017unified,haug2021baselines}, the EWMA has the advantage of reducing the weight of old observations over time. In this way, our baseline automatically adjusts to concept drift.

\subsection{Complexity and Limitations}
The memory complexity of the proposed hierarchical clustering is $\mathcal{O}(K_t w)$, where $w$ is the size of the sliding windows and $K_t$ is the number of nodes at time step $t$. Moreover, the computational complexity of constructing the hierarchical clustering for $T$ data points is $\mathcal{O}(T \log T)$. Accordingly, CDLEEDS has a higher resource consumption than existing methods for global concept drift detection. However, the proposed framework is much more powerful because it can detect both global and local change. 

The selection of an appropriate similarity threshold $\gamma$ is not trivial. If we set $\gamma$ too small, we get large neighborhoods that do not capture local behavior. If we set $\gamma$ too high, the cluster tree may become too deep to be maintained in a real-time application. To address the latter problem, decision tree algorithms often specify a maximum depth. Limiting the depth may result in STNs at leaf nodes that have a lower similarity than originally specified by $\gamma$ (because we cannot further partition the observations). However, to enable more efficient computations, it can often be useful to limit the maximum size of the cluster tree. For example, if we want to use CDLEEDS for global change detection, we do not need the same local granularity as for local change detection.

Similarly, it can be difficult to set a reasonable significance level for hypothesis testing. If the significance level is too small, we might miss certain concept drifts. On the other hand, if the significance level is too high, we might produce many false alarms. However, in our experiments, we obtained good results for common significance levels such as $0.01$ and $0.05$.

Concept drift detection methods are sensitive to hyperparameter settings, and CDLEEDS is no exception. Therefore, it is generally advisable to perform hyperparameter optimization on an initial, stationary training set to learn what degree of variation to expect under a reasonably stable data concept. In addition, it can be useful to re-evaluate the initial hyperparameters at regular intervals.

\begin{table}[t]
\caption{\textit{Data Sets.} We used popular and open-sourced classification data sets in our experiments (obtained from \url{openml.org}, original sources are included where available). TüEyeQ \cite{kasneci2021tueyeq} and Insects \cite{souza2020challenges} comprise natural concept drift. We induced the remaining real-world streaming data sets with artificial concept drift \cite{sethi2017reliable}. Finally, we generated synthetic data streams with scikit-multiflow \cite{montiel2018scikit} (indicated by ``(s.)'').}
    \label{tab:datasets}
    \centering
    \resizebox{\linewidth}{!}{%
        \begin{tabular}{llllll}
        \toprule
        Name & \#Samples & \#Features & \# Classes & Data Types & Drift Types\\
        \cmidrule(lr){1-1} \cmidrule(lr){2-6}
        TüEyeQ \cite{kasneci2021tueyeq} & 15,762 & 77 & 2 & cont., cat. & abrupt \\
        Bank-Marketing \cite{moro2011using} & 45,211 & 16 & 2 & cont., cat. & abrupt\\
        Electricity \cite{harries1999splice} & 45,312 & 8 & 2 & cont., cat. & abrupt\\
        Adult \cite{kohavi1996scaling} &  48,840 & 54 & 2 & cont., cat. & abrupt\\
        Airlines & 539,383 & 7 & 2 & cont., cat. & abrupt\\
        KDD Cup 1999 & 494,020 & 41 & 23 & cont., cat. & abrupt\\
        Covertype \cite{blackard1999comparative} & 581,012 & 54 & 7 & cont., cat. & abrupt\\
        Insects \cite{souza2020challenges} & 355,275 & 33 & 6 & cont. & abrupt\\
        SEA (s.) & 500,000 & 3 & 2 & cont. & abrupt\\
        Agrawal-Gradual (s.) & 500,000 & 9 & 2 & cont. & gradual\\
        Agrawal-Mixed (s.) & 500,000 & 9 & 2 & cont. & abrupt, gradual\\
        \bottomrule
        \end{tabular}
    }
\end{table}

\section{Experiments}\label{sec:experiments}
We evaluated the proposed framework in three experiments. In Section \ref{sec:exp_adaptability}, we show that the proposed hierarchical clustering algorithm is able to identify meaningful spatiotemporal neighborhoods and adapt to local virtual concept drift. In Section \ref{sec:exp_local_drift}, we demonstrate that CDLEEDS can be used to reduce the number of recalculations of local attributions over time. In this context, we also illustrate the local change detection of CDLEEDS. Finally, in Section \ref{sec:exp_global_drift}, we compare CDLEEDS to state-of-the-art methods for global concept drift detection. For illustration, we used a binary and multi-class classification setting, which is well handled by most drift detectors. If not mentioned otherwise, we trained a Hoeffding Tree with adaptive Na\"ive Bayes models at the leaf nodes in the default configuration of scikit-multiflow \cite{montiel2018scikit}. All models and experiments were implemented in Python (3.8.5) and run on an AMD Ryzen Threadripper 3960X CPU with 128GB RAM under Ubuntu 18.04. 

\subsection{Data Sets}
Unfortunately, there are few real-world data sets with known concept drift. Therefore, one usually has to rely on synthetically generated streaming data to evaluate concept drift detection approaches \cite{haug2022standardized}. In our experiments, we used a mixture of popular real-world data sets with both natural and synthetic concept drift, as well as synthetic streaming data sets. We normalized all data sets before use. A list of the data sets and their properties is shown in Table \ref{tab:datasets}.

One of the few real-world data sets with natural and known concept drift is TüEyeQ \cite{kasneci2021tueyeq}, which we already mentioned in the introduction. Recently, \citet{souza2020challenges} presented several data sets with sensor measurements of flying insect species. The classification task is to identify the correct insect. By changing environmental parameters such as humidity and temperature, \citet{souza2020challenges} produced different types of concept drift. In our experiment, we used the unbalanced Insect data set with abrupt concept drift.

Moreover, we imputed popular real-world streaming data sets obtained from \url{openml.org} with synthetic concept drift. In particular, we adopted the method due to \citet{sethi2017reliable} based on the Mutual Information. Specifically, in order to simulate concept drift, we randomly permuted the values of the top 50\% of features with highest Mutual Information with the target. We repeated the procedure to generate multiple synthetic drifts per data set.

We also generated synthetic data streams using scikit-multiflow \cite{montiel2018scikit}. In particular, we generated data streams with abrupt concept drift (SEA), gradual concept drift (Agrawal-Gradual), and mixed, i.e., abrupt and gradual, concept drift (Agrawal-Mixed). We did not balance the classes of the generated data sets and specified \textit{perturbation=0.1} for all generators. Otherwise, we used the default configuration of scikit-multiflow.

\subsection{Hyperparameters for CDLEEDS}
We performed a grid search on the Bank-Marketing data set to identify hyperparameters for CDLEEDS. To obtain unbiased results, we used the same set of hyperparameters in all experiments. Specifically, we set the similarity threshold to $\gamma = 0.95$, the significance level of the t-test to $\alpha = 0.01$, the decay factor of the EWMA-baseline to $\beta = 0.001$, the maximum size of the sliding windows (STNs) to 200 observations, and the maximum age of a node to 100 observations before pruning. If not mentioned otherwise, we limited the depth of the hierarchical clustering to 5. 

\begin{figure}[t]
\centering
\subfloat[Bank-Marketing]{
    \begin{tabular}[b]{c}
        \includegraphics[width=0.435\linewidth]{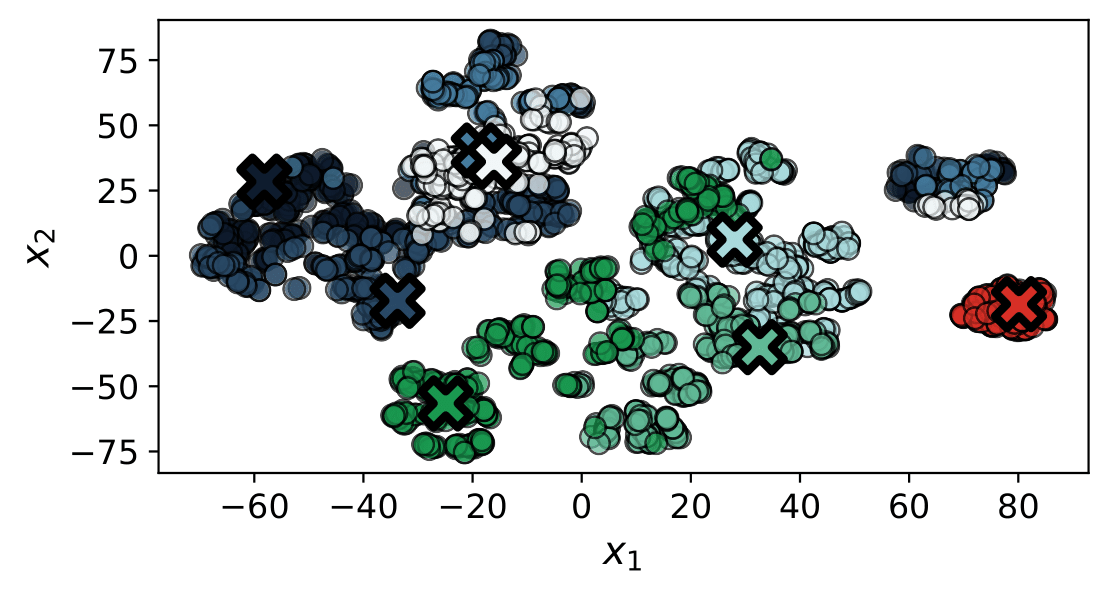}\\
        \includegraphics[width=0.435\linewidth]{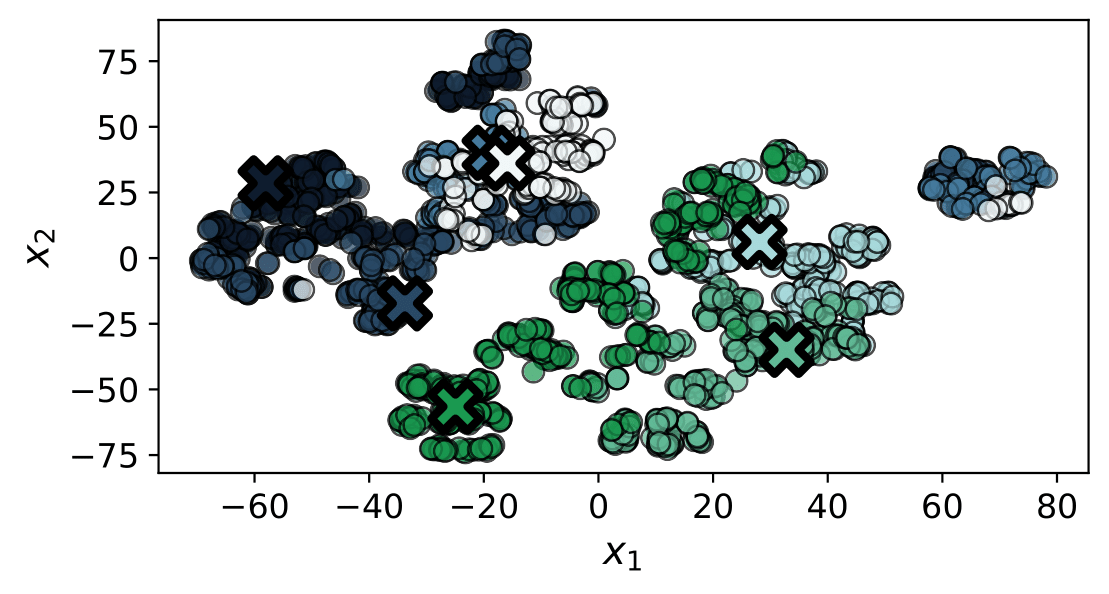}
    \end{tabular}
}
\hfill
\subfloat[Airlines]{
    \begin{tabular}[b]{c}
        \includegraphics[width=0.435\linewidth]{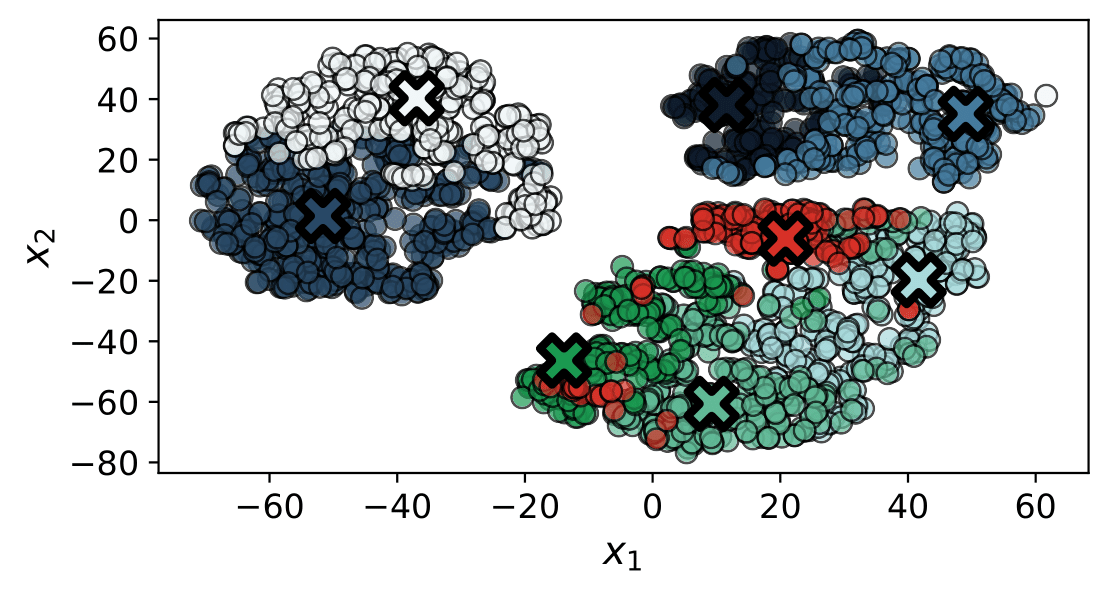}\\
        \includegraphics[width=0.435\linewidth]{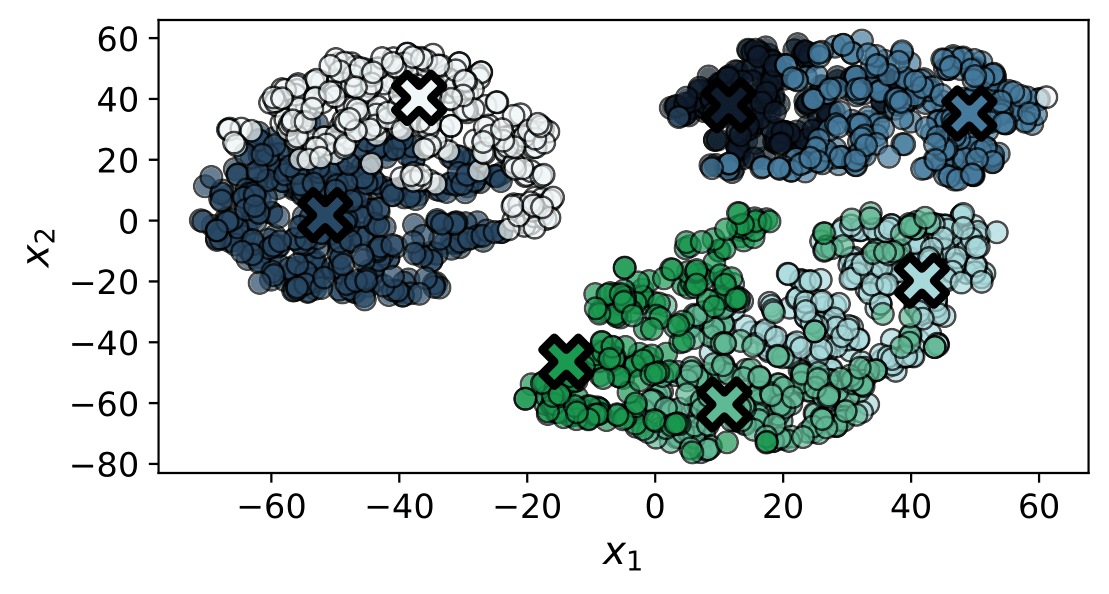}
    \end{tabular}
}
\caption{\textit{CDLEEDS Clustering - Adjusting to Local Virtual Concept Drift.} We trained CDLEEDS on 4,000 observations from two real-world data streams. After processing the first 2,000 observations, we simulated a local virtual concept drift by ignoring all observations that would have fallen into the red cluster during the rest of the training process. Above, we depict the TSNE representation \cite{van2008visualizing} of observations before (upper) and after (lower) the concept drift. The learned clusters are indicated by different colors. The centroids are shown as corresponding ``x'' markers. Notably, the proposed clustering method was able to learn meaningful (i.e., spatially coherent) clusters for both data sets over time.}
\label{fig:local_drift}
\end{figure}

\subsection{1st Experiment - CDLEEDS Clustering Under Local Virtual Concept Drift}\label{sec:exp_adaptability}
In a first experiment, we investigated the ability of the hierarchical clustering method to adapt to local virtual concept drift. Figure \ref{fig:local_drift} shows the TSNE representation of the clustering for two exemplary data sets. Specifically, we collected samples over two time intervals (upper/lower plots). After the first time interval, we simulated a local virtual concept drift. That is, we ignored all new observations that would have been assigned to the red cluster and continued the online training without these observations. In this way, we tested the ability of CDLEEDS to identify and prune obsolete leaves and branches. We limited the maximum depth of the cluster tree to 3 for this experiment. Notably, Figure \ref{fig:local_drift} shows that CDLEEDS managed to form meaningful clusters over time. Moreover, our age-based pruning strategy was able to correctly identify the obsolete (red) cluster. We observed similar results for all remaining data sets.

\subsection{2nd Experiment - Local Change Detection for More Efficient Feature Attributions}\label{sec:exp_local_drift}
\begin{figure}[t]
\centering
\subfloat[Electricity]{
    \includegraphics[width=.48\linewidth]{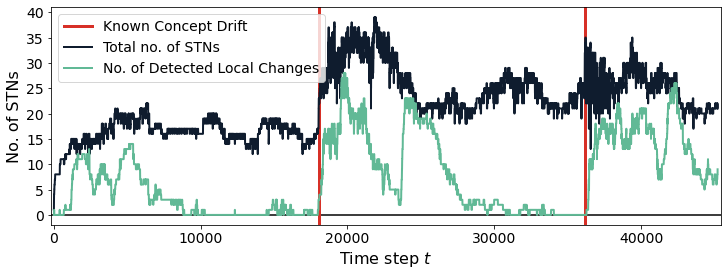}
}
\hfill
\subfloat[Bank-Marketing]{
    \includegraphics[width=.48\linewidth]{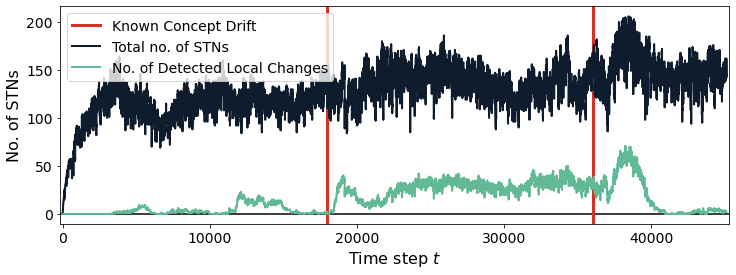}
}
\hfill
\subfloat[Adult]{
    \includegraphics[width=.48\linewidth]{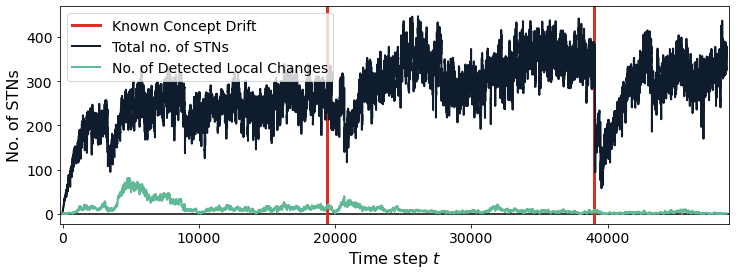}
}
\hfill
\subfloat[Airlines]{
    \includegraphics[width=.48\linewidth]{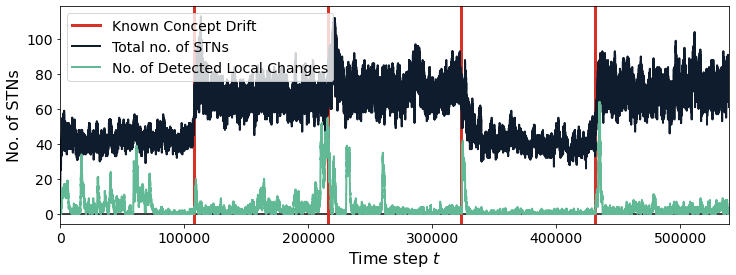}
}
\caption{\textit{Local Change Detection with CDLEEDS.} We show the CDLEEDS clustering on 4 data sets. In particular, we depict the total number of STNs over time (blue) and the number of STNs for which we detected local change (green). The red vertical lines indicate known global concept drifts. In all cases, the CDLEEDS clustering changed in complexity after a concept drift and/or detected an increasing number of local changes. Yet, concept drift often only affected a subset of the STNs. Moreover, in times of stable data concepts, there were usually only a few STNs underlying changes (caused by the continued incremental model updates). With this insight, we could considerably reduce the number of local attributions that have to be recalculated after each update.}
\label{fig:leaves_drift}
\end{figure}

CDLEEDS is a local change detection framework that can help make local attribution methods in data streams more feasible. In this experiment, we demonstrate the ability of the proposed framework to detect local changes, in particular those caused by concept drift. Figure \ref{fig:leaves_drift} illustrates the number of spatiotemporal $\gamma$-neighborhoods, i.e., leaf nodes, maintained by CDLEEDS for four exemplary data sets. We also show the number of detected local changes over time. We did not limit the maximum depth of the hierarchical clustering for this experiment. Consequently, at each time step, all leaf nodes corresponded to valid STNs with $\gamma = 0.95$ as defined in Def. \ref{def:stn}.

\begin{figure}[t]
\centering
\captionsetup{width=.48\linewidth} 
\subfloat[Adult - Reduction of recalc.: 92.82\% upper plot, 62.40\% lower plot, 80.05\% $\pm$ 0.53\% on average (100 obs.).]{
    \begin{tabular}[b]{@{}c@{}}
        \includegraphics[width=0.48\linewidth]{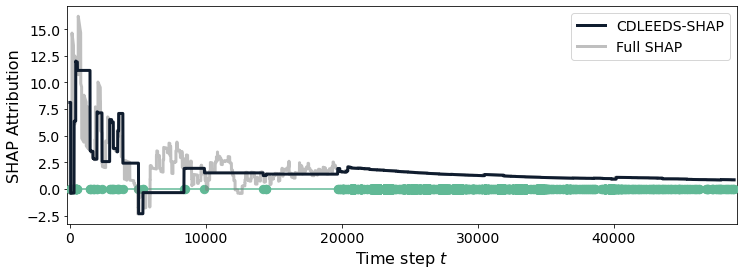}\\
        \includegraphics[width=0.48\linewidth]{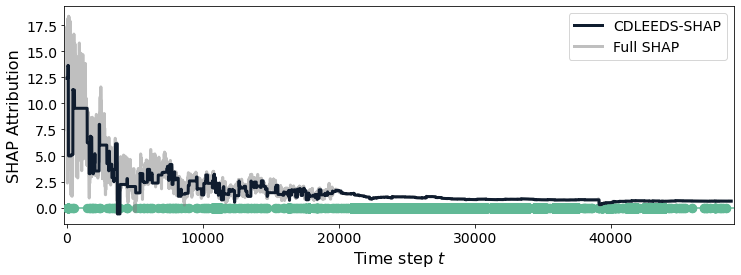}
    \end{tabular}
}
\hfill
\subfloat[Bank-Mark. - Reduction of recalc.: 99.85\% upper plot, 88.77\% lower plot, 95.71\% $\pm$ 0.05\% on average (100 obs.).]{
    \begin{tabular}[b]{@{}c@{}}
        \includegraphics[width=0.48\linewidth]{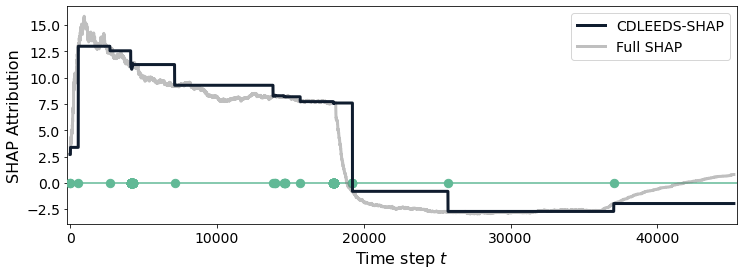}\\
        \includegraphics[width=0.48\linewidth]{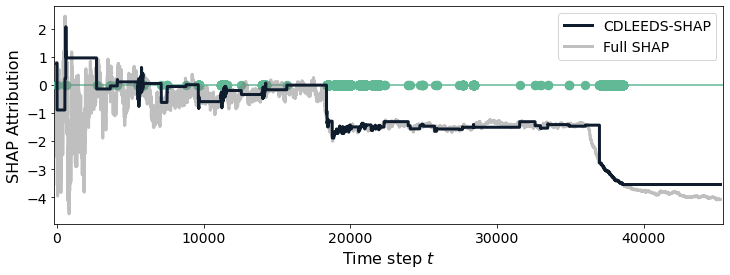}
    \end{tabular}
}
\hfill
\subfloat[TüEyeQ - Reduction of recalc.: 99.96\% upper plot, 89.90\% lower plot, 95.97\% $\pm$ 0.12\% on average (100 obs.).]{
    \begin{tabular}[b]{@{}c@{}}
        \includegraphics[width=0.48\linewidth]{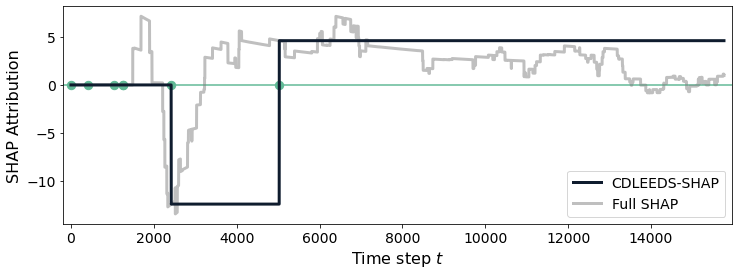}\\
        \includegraphics[width=0.48\linewidth]{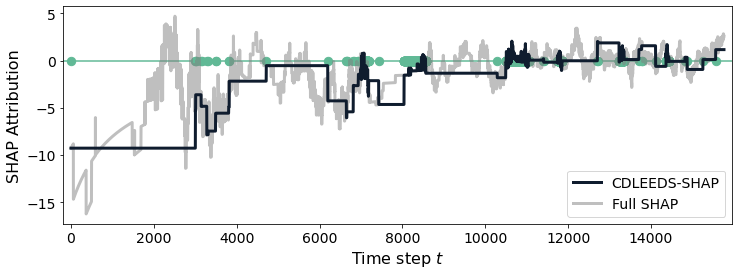}
    \end{tabular}
}
\hfill
\subfloat[Electricity - Reduction of recalc.: 70.26\% upper plot, 51.23\% lower plot,  56.11\% $\pm$ 0.11\% on average (100 obs.).]{
    \begin{tabular}[b]{@{}c@{}}
        \includegraphics[width=0.48\linewidth]{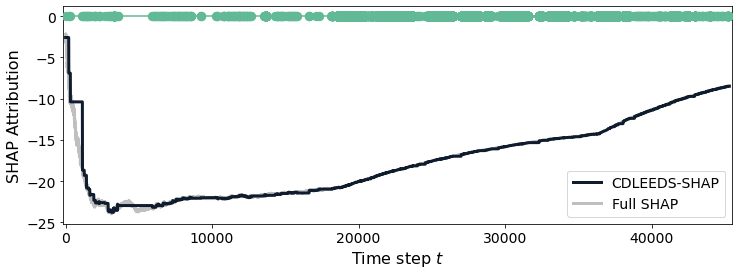}\\
        \includegraphics[width=0.48\linewidth]{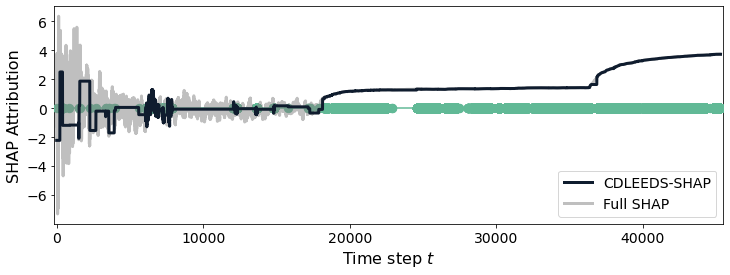}
    \end{tabular}
}
\captionsetup{width=\linewidth} 
\caption{\textit{Feasible Local Attributions in Data Streams with CDLEEDS.} From a random sample of 100 observations per data set, we show the SHAP attributions \cite{lundberg2017unified} of the two observations that required the fewest (top) and most (bottom) recalculations over time. For reasons of readability, we only display the attribution of the feature with the largest average value. However, note that we observed similar results for all observations and features. The blue line indicates the SHAP attributions that we obtained by using CDLEEDS to trigger recalculations (detected local changes are indicated by green markers). The grey background pattern corresponds to the actual SHAP attribution at each time step. By using CDLEEDS, the average number of recalculations can be significantly reduced without affecting the explanatory power compared to the actual feature attributions.}
\label{fig:shap_recomputations}
\end{figure}

Based on Figure \ref{fig:leaves_drift}, we can make several interesting observations. As in the previous experiment, we find that the hierarchical clustering method is able to adapt to concept drift by pruning obsolete leaves and branches or creating new ones. This adaptation is most evident after the last concept drift in the Adult data set. Notably, during this period, CDLEEDS issued only few local change alerts, suggesting that the last concept drift in Adult is a virtual rather than a real concept drift. Early change detections can generally be attributed to the initial training of the predictive model and cluster tree and would thus be ignored in practice. In general, the known concept drifts are accompanied by a substantial increase in the detected local drifts. Moreover, Figure \ref{fig:leaves_drift} shows that there is usually relatively little local change when the data generating distribution is stable. 

We can use this insight to make the recomputation of local attributions more efficient. In particular, we argue that it is usually sufficient to recompute old attributions when there has been a corresponding local change. To support our argument, we computed SHAP attributions \cite{lundberg2017unified} for four different data streams. This time we used a logistic regression model, as the SHAP implementation for linear models is much more efficient. Also, we only performed this experiment on the small data sets, as calculating and storing SHAP values for large data streams is not feasible on most machines.

We conducted the experiment as follows: At time step $t=0$, we computed the SHAP attributions for a random sample of 100 observations. We then recomputed the SHAP attribution of an observation in subsequent time steps, only if the observation had been assigned to a new leaf in the hierarchical clustering, or if a corresponding local change had been detected. Figure \ref{fig:shap_recomputations} shows SHAP attributions for two observations in each data stream. Strikingly, the CDLEEDS-based recomputations approximate the actual SHAP attributions well. Indeed, for the entire sample of 100 observations, we observed an average deviation from the true SHAP attribution of only 0.10 $\pm$ 0.15 for Adult, 0.16 $\pm$ 0.10 for Bank-Marketing, 0.45 $\pm$ 0.29 for TüEyeQ, and 0.11 $\pm$ 0.11 for Electricity. Given attributions of up to 17.5 (see Adult) or -24 (see Electricity), these deviations become negligible. At the same time, CDLEEDS was able to considerably reduce the number of recalculations. We observed an average reduction in recalculations of 80.05\% $\pm$ 0.53\% for Adult, 95.71\% $\pm$ 0.05\% for Bank-Marketing, 95.97\% $\pm$ 0.12\% for TüEyeQ, and 56.11\% $\pm$ 0.11\% for Electricity (in \% of all time steps). 

Our experiments show that periodic recalculations of local attributions are generally necessary, since attributions can change considerably in the streaming setting. However, the number of recalculations actually performed can be significantly reduced through CDLEEDS. In addition, the detected changes, along with the recalculated attributions, may carry valuable explanatory information. For example, in Figure \ref{fig:shap_recomputations}(c), drastic changes in the local attributions and corresponding alerts by CDLEEDS indicate a sudden concept drift around $t=2500$ (which might be due to a new, e.g., more difficult, type of IQ task \cite{kasneci2021tueyeq}). In general, as claimed above and shown in our experiments, CDLEEDS can make local attribution-based explainability in data streams more efficient and expressive.

\subsection{3rd Experiment - Using CDLEEDS for Global Concept Drift Detection}\label{sec:exp_global_drift}
CDLEEDS is designed as a framework for detecting local change. However, as suggested in Section \ref{sec:cluster}, we might also use CDLEEDS to detect global concept drift by applying a simple strategy based on Fisher's method for combining p-values. For the sake of completeness, we thus compared CDLEEDS with several state-of-the-art global concept drift detection models. In particular, we compared our approach to ADWIN \cite{bifet2007learning}, DDM \cite{gama2004learning}, ECDD \cite{ross2012exponentially}, MDDM-A \cite{pesaranghader2018mcdiarmid} and RDDM \cite{barros2017rddm}, all of which attempt to detect concept drift through changes in the error rate. Moreover, we compared CDLEEDS to ERICS \cite{haug2021learning}, which detects concept drift by monitoring changes in the distributions of model parameters. We used the original ERICS implementation provided by the authors \cite{haug2021learning}. The remaining implementations are openly available via the tornado package \cite{pesaranghader2018reservoir}. We applied the same hyperparameter search as for CDLEEDS. Accordingly, we specified $\text{\textit{delta}}=0.1$ for ADWIN \cite{bifet2007learning}, as well as $\text{\textit{window\_mvg\_average}}=90$ and $\text{\textit{beta}}=0.001$ for ERICS \cite{haug2021learning}. Other than that, however, we could use the default hyperparameters.

\subsubsection{Evaluation Measures}
We examined the delay, recall, and false discovery rate (FDR) of each drift detection method \cite{haug2022standardized}. The delay corresponds to the time steps until a known concept drift is first detected. The recall quantifies the proportion of known concept drifts that the model detected. And the FDR is the proportion of false positives among all detected drifts. Note that only the combination of recall and FDR provides a meaningful evaluation, as each measure can be easily optimized on its own. Therefore, we report the mean of recall and (1-FDR) in Table \ref{tab:recall_fdr}. To compute the recall and FDR, we need to define a time interval after known concept drift in which we count a drift alert as a true positive. In the experiments, we used several intervals with lengths between 1\% and 10\% of the original data set size. Table \ref{tab:recall_fdr} shows the means and standard deviations for the different interval sizes. In addition, to give the classifier time for initial training, we did not include drift alerts that occurred within the first 1,000 observations.

\subsubsection{Results}
In general, we find that there are considerable differences between all concept drift detection methods and data sets, both regarding the combined recall and FDR in Table \ref{tab:recall_fdr} and the delay in Table \ref{tab:delay}. As described above, this effect might be mitigated by (periodically) optimizing the hyperparameter configurations for each data set. However, such performance differences can often be observed in practice because concept drift detection methods are usually sensitive to the predictive model and data distribution at hand. For this reason, it is generally advisable to use multiple methods in parallel for more robust global concept drift detection. 

Naturally, the more elaborate drift detectors CDLEEDS (2.32 milliseconds) and ERICS \cite{haug2021learning} (3.68 ms) had a larger average update time than the error rate-based drift detectors (ADWIN \cite{bifet2007learning} = 0.14 ms, DDM \cite{gama2004learning} = 0.12 ms, ECDD \cite{ross2012exponentially} = 0.12 ms, MDDM-A \cite{pesaranghader2018mcdiarmid} = 0.14 ms, and RDDM \cite{barros2017rddm} = 0.12 ms). However, the computation times should be treated with care, as they generally depend on the implementation and hardware configuration at hand.

Despite its relative simplicity, the CDLEEDS-based approach to global concept drift detection competes with powerful existing methods such as ADWIN \cite{bifet2007learning}, ECDD \cite{ross2012exponentially} or ERICS \cite{haug2021learning}. In particular, CDLEEDS received the best average score and the second best average ranking on the combined recall and FDR measure. At the same time, CDLEEDS was usually also able to achieve a short delay. In summary, our results suggest that CDLEEDS, although being originally designed for local change detection, might also be a valuable alternative to popular global drift detection methods.

\begin{table}[t]
\caption{\textit{Global Concept Drift Detection -- Part 1.} CDLEEDS was developed for local change detection, in particular to detect obsolete local attributions. However, as a byproduct, CDLEEDS can also be used to detect global concept drift. The results of this additional experiment are shown in this and the following table. The missing values correspond to test runs in which a method did not raise an alert. ERICS \cite{haug2021learning} can only process binary target variables, so no results are available for the multi-class data sets. Here we show the mean of recall + (1 - false discovery rate) (\%; mean $\pm$ standard deviation; higher is better). Strikingly, CDLEEDS can compete with state-of-the-art methods for concept drift detection.}
    \label{tab:recall_fdr}
    \centering
    \resizebox{\linewidth}{!}{%
        \begin{tabular}{llllllll}
        \toprule
        {} & CDLEEDS & ERICS & ADWIN &	DDM & 	ECDD & MDDM-A &	RDDM \\
        \midrule
        TüEyeQ &	0.57	$\pm$ 	0.04 &	0.61	$\pm$ 	0.04 &	0.50	$\pm$ 	0.17 &	0.38	$\pm$ 	0.16 &	0.57	$\pm$ 	0.18 &	0.28	$\pm$ 	0.17 &	0.36	$\pm$ 	0.26\\
        Bank-Mark. &	0.58	$\pm$ 	0.01 &	0.56	$\pm$ 	0.03 &	0.45	$\pm$ 	0.00 & 	- &	0.56	$\pm$ 	0.03 &	0.75	$\pm$ 	0.00 &	0.63	$\pm$ 	0.00\\
        Electricity &	0.40	$\pm$ 	0.00 &	0.56	$\pm$ 	0.02 &	0.45	$\pm$ 	0.06 &	-	& 0.26	$\pm$ 	0.00 &	0.00	$\pm$ 	0.00 &	0.25	$\pm$ 	0.00\\
        Adult &	0.66	$\pm$ 	0.04 &	0.57	$\pm$ 	0.03 &	0.41	$\pm$ 	0.24 &	0.25	$\pm$ 	0.00 &	0.56	$\pm$ 	0.02 &	0.00	$\pm$ 	0.00 &	0.07	$\pm$ 	0.00\\
        Airlines &	0.62	$\pm$ 	0.06 &	0.61	$\pm$ 	0.06 &	0.70	$\pm$ 	0.07 &	0.63	$\pm$ 	0.00 &	0.57	$\pm$ 	0.09 &	0.37	$\pm$ 	0.05 &	0.35	$\pm$ 	0.00\\
        SEA &	0.60	$\pm$ 	0.09 &	0.19	$\pm$ 	0.01 &	0.65	$\pm$ 	0.05 &	0.85	$\pm$ 	0.09 &	0.60	$\pm$ 	0.05 &	0.78	$\pm$ 	0.08 &	0.83	$\pm$ 	0.03\\
        Agrawal-Grad. &	0.68	$\pm$ 	0.02&	0.70	$\pm$ 	0.03 &	0.75	$\pm$ 	0.02 &	0.42	$\pm$ 	0.00 &	0.70	$\pm$ 	0.02 &	0.67	$\pm$ 	0.00 &	0.32	$\pm$ 	0.01\\
        Agrawal-Mix. &	0,55	$\pm$ 	0.12 &	0.62	$\pm$ 	0.04 &	0.69	$\pm$ 	0.04 &	0.55	$\pm$ 	0.00 &	0.63	$\pm$ 	0.04 &	0.65	$\pm$ 	0.06 &	0.40	$\pm$ 	0.00\\
        KDD Cup &	0.59	$\pm$ 	0.13 & 	- &	0.45	$\pm$ 	0.21 &	0.12	$\pm$ 	0.00 &	0.61	$\pm$ 	0.05 &	- &	0.40	$\pm$ 	0.15\\
        Covertype &	0.49	$\pm$ 	0.08 & 	- &	0.49	$\pm$ 	0.02 &	0.33	$\pm$ 	0.00 &	0.61	$\pm$ 	0.06 &	0.28	$\pm$ 	0.00 &	0.47	$\pm$ 	0.05\\
        Insects &	0.62	$\pm$ 	0.06 &	-	& 0.68	$\pm$ 	0.11 &	- &	0.51	$\pm$ 	0.08 &	0.62	$\pm$ 	0.08 &	0.52	$\pm$ 	0.08\\
        \midrule
        Mean Measure &	0.58	$\pm$ 	0.06 &	0.55	$\pm$ 	0.03 &	0.56	$\pm$ 	0.09 &	0.44	$\pm$ 	0.03 &	0.56	$\pm$ 	0.06 &	0.44	$\pm$ 	0.05 &	0.42	$\pm$ 	0.05\\
        Mean Ranking & 2.9 &	3.1 &	2.6 &	4.1 &	3.2 &	4.4 &	4.8 \\
        \bottomrule
        \end{tabular}
    }
\end{table}

\begin{table}[t]
\caption{\textit{Global Concept Drift Detection -- Part 2.} Below we depict the drift detection delay (no. of observations; lower is better). As in the previous table, missing values correspond to test runs in which a method did not issue a single alert.}
    \label{tab:delay}
    \centering
    \resizebox{.9\linewidth}{!}{%
        \begin{tabular}{llllllll}
        \toprule
        {} & CDLEEDS & ERICS & ADWIN &	DDM & 	ECDD & MDDM-A &	RDDM \\
        \midrule
        TüEyeQ & 940 &	39 &	357 &	1,049 &	282 &	973 &	819 \\
        Bank-Mark. &	238 &	7 &	4,586 &	- &	181 &	4,582 &	4721 \\
        Electricity &	5,859 &	98 &	4,656 &	- &	9,097 &	13,643 &	13,643\\
        Adult &	204 &	2 &	1,851 &	14,701 &	60 &	14,701 &	14,702\\
        Airlines &	819 &	20 &	144 &	81,262 &	4,482 &	54,043 &	54,035\\
        SEA &	2,517 &	75,032 &	675 &	27,959 &	76 &	25,897 &	1,410\\
        Agrawal-Grad. &	13,251 &	30 &	5,987 &	100,415 &	328 &	12,869 &	101,306\\
        Agrawal-Mix. &	10,163 &	27 &	1,315 &	36,515 &	173 &	2,417 &	33,691\\
        KDD Cup &	5,412 &	- &	20,177 &	98,829 &	550 &	- &	25,312\\
        Covertype &	11,316 &	- &	27,817 &	116,227 &	366 &	58,408 &	14,878\\
        Insects &	152 &	- &	4,160 &	- &	6,091 &	2,050 &	17,552\\
        \midrule
        Mean Ranking &	3.3 &	1.8 &	3.1 &	6.1 &	2.3 &	4.6 &	5.0\\
        \bottomrule
        \end{tabular}
    }
\end{table}

\section{Conclusion}
To the best of our knowledge, this is the first work to formally investigate the mechanisms underlying changes in local feature attributions in data streams. It turns out that for sensible attribution methods that respect the local accuracy criterion, attribution changes are a direct consequence of incremental model updates and concept drift. CDLEEDS, the framework proposed in this work, can reliably detect such changes both locally and globally, thereby enabling more efficient and reliable use of attribution methods in online machine learning. Indeed, the proposed framework can compete with state-of-the-art concept drift detection methods, which we have demonstrated in extensive experiments on publicly available data sets. Accordingly, CDLEEDS is a flexible tool that enables more efficient, robust, and meaningful explainability in data stream applications.

\bibliographystyle{ACM-Reference-Format}


\end{document}